# Improved Learning of Bayesian Networks

**Tomáš Kočka**
Laboratory for Intelligent Systems
Univ. of Economics Prague
Czech Republic
kocka@vse.cz

**Robert Castelo**
Institute of Information & Computing Sciences
University of Utrecht
The Netherlands
roberto@cs.uu.nl

## Abstract

The search space of Bayesian Network structures is usually defined as Acyclic Directed Graphs (DAGs) and the search is done by local transformations of DAGs. But the space of Bayesian Networks is ordered with respect to inclusion and it is natural to consider that a good search policy should take this into account. The first attempt to do this (Chickering 1996) was using equivalence classes of DAGs instead of DAGs itself. This approach produces better results but it is significantly slower. We present a compromise between these two approaches. It uses DAGs to search the space in such a way that the ordering by inclusion is taken into account. This is achieved by repetitive usage of local moves within each equivalence class of DAGs. We show that this new approach produces better results than the original DAGs approach without substantial change in time complexity. We present empirical results, within the framework of heuristic search and Markov Chain Monte Carlo, provided through the Alarm dataset.

## 1 Introduction

A Bayesian Network $G$ for a set of variables $V = \{x_1, ..., x_n\}$ represents a joint probability distribution over those variables. It consists of (I) a network structure that encodes assertions of conditional independence in the distribution and (II) a set of local conditional probability distributions corresponding to that structure. The network structure is an acyclic directed graph (DAG) such that each node corresponds to one variable in $V$.

There are many methods for learning Bayesian Networks from data. They usually consist of three components: (1) a *search space* and a *traversal operator* that by means of local transformations of the structure defines a set of *neighbours*, (2) a *scoring metric* evaluating the quality of a given structure and (3) a *search strategy*.

The basic problem is the choice of the search space, because there exists different DAGs which assert the same set of independence assumptions among the variables in the domain - we call such networks *equivalent*. So the problem is whether to search the *space of all DAGs* or the *space of all equivalence classes of DAGs* (*essential graphs* or another representation).

The advantage of the space of all equivalence classes is that it is smaller than the space of all DAGs. On the other hand it is not possible to compare the scores of two essential graphs by local computations derived directly from their graph structure. One should first transform those essential graphs into DAGs in order to do so. Thus, it becomes difficult to define the traversal operator as it needs to use non-local algorithms converting DAG to essential graph and essential graph to DAG which makes it computationally expensive. For DAGs the simple alternative of add, remove and reverse an arc is often used as the traversal operator. It was shown in (Chickering 1996) that the learning via equivalence classes produces better results than the DAGs approach but it needs significantly more time.

In the context of the Markov Chain Monte Carlo (MCMC) method, Madigan *et al.* (1996) show that in order to build an irreducible Markov chain over the space of essential graphs it is necessary to design a traversal operator that modifies two edges at once.

In this paper we argue that the better results presented by (Chickering 1996) are not caused by the usage of equivalence classes only but by the combination of the space of equivalence classes and the traversal operator they used. Moreover, recent results of (Gillispie 2001) and (Steinsky 2000) suggest that the space of all equivalence classes is only cca 3.7 times smaller than the space of all DAGs. So the use of equivalence classes does not provide a substantial reduction with respect of the size of the search space of DAGs. We show that the space of DAGs (which is computationally cheaper to use) can produce similar results and much faster.

There are two major contributions of this paper. First, we introduce a new concept of traversal operator for the DAGs space. This concept is based upon the (Chickering 1995) transformational characterization of equivalent DAGs. It uses transformations among equivalent DAGs instead of equivalence classes and so all necessary operations are local. It is sensible to expect that learning algorithms that consider the inclusion among Bayesian Networks will perform better than those that do not. The approach we use is based upon the result of (Kočka et al. 2001) characterizing the inclusion of DAGs that differ in at most one adjacency. It is enough for us as long as we want to use only local changes to the structures, but the general inclusion is still an open problem.

The second contribution of this paper is the implementation of the previous idea within the frameworks of *heuristic learning* and the MCMC method. Our experiments show that this approach produces better results than the standard operator for the DAGs space without substantial change in time complexity. The MCMC implementation not only will be an improvement by itself but it will help understanding why the approach achieves such an improvement. The experiments have been carried out using the standard benchmark dataset of the Alarm network used previously by Cooper and Herskovits (1992) and in many other works within the subject.

In the next section we introduce the basic concepts of DAGs, their equivalence classes, their inclusion and a brief comment about the sizes of DAG and essential graph spaces. In the section 3 we formalize the different concepts of neighbourhood, and provide their implementation in the framework of heuristic search and MCMC. These neighbourhoods, within the particular implementation we provide, will be compared in section 4 using the well known benchmark Alarm dataset. We end with concluding remarks in section 5.

## 2 Basic concepts

In this section notation and previous relevant work to this paper are introduced.

Lower case letters are used to denote elements of $V$ while capital letters are used to denote subsets of $V$. Possibly indexed capital letters $L, G, H$ will be used to denote DAGs over $V$. We use $\mathcal{E}(G)$ for the underlying (undirected) *skeleton* of the DAG $G$.

The symbol $\langle A, B|C \rangle$ denotes a triplet of pairwise disjoint subsets $A, B, C$ of $V$. The symbol $\mathcal{T}(V)$ will denote the class of all disjoint triplets over $V$:

$$\{\langle A, B|C\rangle;\ A, B, C \subseteq V\ A \cap B = B \cap C = A \cap C = \emptyset\}$$

The induced independence model $\mathcal{I}(G)$ is defined as follows:

$$\mathcal{I}(G) = \{\,\langle A, B|C\rangle \in \mathcal{T}(V)\,;\quad A \perp\!\!\!\perp B \mid C\ [G]\,\}$$

where $A \perp\!\!\!\perp B \mid C\ [G]$ means that this independence is valid in the DAG $G$ according to the d-separation criterion.

### 2.1 Equivalence of DAGs

The following definition is extremely relevant to the specification of the new traversal operator we will introduce later.

DEFINITION 2.1 An edge $a \to b$ in $G$ is *covered in $G$* if $pa_G(a) \cup a = pa_G(b)$.

LEMMA 2.1 Supposing $G$ and $H$ are DAGs over $V$. Then, the following three conditions are equivalent

(1) $\mathcal{I}(G) = \mathcal{I}(H)$ ($G$ and $H$ are equivalent DAGs).

(2) $\mathcal{E}(G) = \mathcal{E}(H)$ and the graphs $G$ and $H$ have the same immoralities.

(3) there exists a sequence $L_1, \ldots, L_m$, $m \geq 1$ of DAGs over $V$ such that $L_1 = G, L_m = H$ and $L_{i+1}$ is obtained from $L_i$ by reversing an arc which is covered in $L_i$ for $i = 1, \ldots, m-1$.

The equivalence (1) $\Leftrightarrow$ (2) was proved in (Verma and Pearl 1991), in more general framework of chain graphs in (Frydenberg 1990); the equivalence (1) $\Leftrightarrow$ (3) was proved in (Chickering 1995) and (Heckerman et al. 1995).

### 2.2 Inclusion of DAGs

We say that the independence model of a DAG $G$ *is included* in the independence model of a DAG $H$ when $\mathcal{I}(G) \subseteq \mathcal{I}(H)$. The concept of inclusion defines a natural ordering of the space of Bayesian Networks. The following conjecture has not been proven in its general form yet. However no counterexample is known against it till these days.

CONJECTURE 1 (Meek 1997)
Let $G$ and $H$ be DAGs over a set of variables $N$. Then $\mathcal{I}(G) \subseteq \mathcal{I}(H)$ iff there exists a sequence of DAGs $L_1, \ldots, L_n$ such that $G = L_1$, $H = L_n$ and the graph $L_{i+1}$ is obtained from $L_i$ by applying either the operation of covered arc reversal or the operation of arc removal for $i = 1, \ldots, n - 1$.

It is easy to show that the existence of the sequence implies $\mathcal{I}(G) \subseteq \mathcal{I}(H)$. But a special case of the conjecture has been recently proved:

LEMMA 2.2 (Kočka et al. 2001)
Let $G$ and $H$ be DAGs over a set of variables $V$. If $\mathcal{I}(G) \subseteq \mathcal{I}(H)$ and $|\mathcal{E}(G)| = |\mathcal{E}(H)| + 1$ then there exists a sequence of DAGs $L_1, \ldots, L_n$ such that $G = L_1$, $H = L_n$ and the graph $L_{i+1}$ is obtained from $L_i$ by applying the operation of covered arc reversal for $i = 1, \ldots, j - 1$, the operation of arc removal for $i = j$ and the operation of covered arc reversal for $i = j + 1, \ldots, n - 1$ where $j \in \{1, \ldots, n - 1\}$.

Let $H, K, L$ denote DAGs. Let $\mathcal{I}(K) \prec \mathcal{I}(L)$ denote $\mathcal{I}(K) \subset \mathcal{I}(L)$ and for no $H$, $\mathcal{I}(K) \subset \mathcal{I}(H) \subset \mathcal{I}(L)$. The *inclusion boundary* of the induced independence model $\mathcal{I}(G)$ of a DAG $G$ is defined as

$$\mathcal{IB}(G) = \{\mathcal{I}(K); \mathcal{I}(K) \prec \mathcal{I}(G)\} \cup \{\mathcal{I}(L); \mathcal{I}(G) \prec \mathcal{I}(L)\}$$

Let $\mathbf{C} = \{G_1, \ldots, G_n\}$ be the set of equivalent DAGs that forms any given equivalence class of Bayesian Networks. From lemmas 2.2 and 2.1 it follows that any independence model $\mathcal{I}(H)$, induced by every DAG $H$ obtained by adding or removing an arc from any $G_i \in \mathbf{C}$, is contained in $\mathcal{IB}(G)$.

### 2.3 Complexity of the search space

From the asymptotic for the number of DAGs given by Robinson (1973) it follows that the search space of DAGs grows more than exponentially. One may think that the number of equivalence classes of DAGs (represented by essential graphs) departs substantially from such rate, but recent empirical investigation gives a quite different perspective.

OBSERVATION 1 (Gillispie 2001)
*number of DAGs/number of essential graphs* $< 3.7$ was confirmed up to 9 nodes.

Moreover (Steinsky 2000) shows that the ratio has some limit (smaller than 14) for up to 200 nodes. This result makes the previous observation very probable to hold in general.

## 3 The RCAR neighbourhood

The most rigorous operation for a Random Equivalent DAG Selection (REDS) for given DAG should be based upon the proof of lemma 2.1 (3) in (Chickering 1995).

From the proof it follows that to convert one DAG to another equivalent DAG it is not necessary to reverse one arc more than once. This was the motivation for the following algorithm:

**The REDS algorithm**

1. Let $r_i$ be a random number assigned to each vertex $x_i \in V$ for a given DAG $G = (V, E)$.

2. Let $\mathcal{C} = \{e \in E | e \text{ is covered}\}$. Let $\mathcal{R} = \{x_j \to x_k \in \mathcal{C} | r_j < r_k\}$.

3. If $\mathcal{R} = \emptyset$ then end, otherwise select randomly an arc $e \in \mathcal{R}$, reverse it, print $G$ and go to step 2.

This algorithm outputs a sequence of equivalent DAGs $G_1, \ldots, G_m$ where $G_m$ has a local maximum of arcs oriented according to the random ordering of the nodes. All equivalent DAGs are possible thanks to the lemma 2.1 but the question if they are equally probable remains open.

Another algorithm which will produce any equivalent DAG to a given DAG with nonzero probability is the Repeated Covered Arc Reversal (RCAR). We use this algorithm as it is less computationally demanding than the REDS algorithm. The RCAR algorithm has one free parameter $r$ which stands for an upper bound of the number of repetitions. We propose to use some small constant like 4 or 10 which should be enough for most of DAGs thanks to the observation 1 .

**The RCAR algorithm**

1. Generate a random number $rr$ between 0 and $r$. Repeat the following two steps $rr$ times.

2. Search for all covered arcs in the DAG $G$.

3. Select at random one arc from the set of all covered arcs in $G$ and reverse it.

### 3.1 Concepts of neighbourhood

Now we have the RCAR operation and we can summarize the most often used concept of neighbourhood (AR), some of its restrictions and its generalizations using the RCAR operation. The concepts we assume in our experiments are:

- **NR** (No Reversals) All DAGs with one arc more and one arc less that do not introduce a directed cycle.

- **AR** (All Reversals) The NR neighbourhood plus all DAGs with one arc reversed that does not introduce a directed cycle.

- **CR** (Covered Reversals) The NR neighbourhood plus all DAGs with one *covered* arc reversed.
- **NCR** (Non-Covered Reversal) The NR neighbourhood plus all DAGs with one *non-covered* arc reversed.
- **RCARR** (RCAR+Reversals) Perform the RCAR algorithm and then create a NCR neighbourhood.
- **RCARNR** (RCAR+NR) Perform the RCAR algorithm and then create a NR neighbourhood.

In order to avoid local maxima as much as possible, *ideally* a search strategy should let an independence model $\mathcal{I}(G)$, reach any other independence model in its inclusion boundary $\mathcal{IB}(G)$, in one single step. It is clear that the NR, CR, AR and NCR neighbourhoods do not satisfy this property. However, the RCARR or RCARNR neighbourhoods do satisfy this property as long as the number of repetitions within RCAR is large enough and Meek's conjecture holds. Even in the case when one of these two last conditions does not hold (or both), the RCARR and RCARNR neighbourhoods clearly allow to reach much more members of the inclusion boundary than the other traditional neighbourhoods.

### 3.2 Heuristic learning

We are dealing with the general learning problem where no causal ordering is assumed. The simplest, and probably most used, heuristic learning algorithm for Bayesian Networks has been a *hill-climber* (greedy search) with a traversal operator that creates an AR neighbourhood.

```
algorithm hcmc(int r, bool ncr) returns dag
  dag g = emptydag
  bool local_maxima = false
  int trials = 0
  while (¬local_maxima) do
    g.rcar(r)
    set nh = g.neighbourhood(ncr)
    dag g' = g.score_and_pick_best(nh)
    local_maxima = (g'.score() < g.score())
    if (¬local_maxima) then
      g = g'
      trials = 0
    else if (trials < MAXTRIALS) then
      g.rcar(r)
      local_maxima = false
      trials = trials + 1
    endif
  endwhile
  return g
endalgorithm
```

Figure 1: Hill-Climber Monte Carlo algorithm

We propose two simple modifications of the hill-climber. Firstly, we want that the traversal operator is preceded by a RCAR operation, creating a RCARR or a RCARNR neighbourhood. Secondly, we will try, in a limited number of trials, to escape from local maxima by performing RCAR. We may see the implementation of these two modifications in the algorithm in figure 1, which we will call Hill-Climber Monte Carlo (HCMC) in the rest of the paper due to its random nature. We have used both these modifications together and we believe that this combination makes most sense.

Moreover we use two alternatives for the neighbour operator. Either it performs no reversals at all (NR), or only non-covered reversals (NCR).

### 3.3 MCMC learning

The MCMC method samples from a target distribution, in this case the posterior distribution of Bayesian Networks given the data $p(G|D)$. Madigan and York (1995) adapted the Metropolis-Hastings sampler for Bayesian and Decomposable Networks, providing the so-called Markov Chain Monte Carlo Model Composition, commonly known as $MC^3$, which implements an aperiodic and irreducible Markov chain that moves in the search space of graphs with stationary distribution $p(G|D)$. Given a DAG $G$, the $MC^3$ algorithm *proposes* a move which consists of a local transformation of $G$ that leads to a randomly chosen candidate DAG $G'$. This proposed move will be accepted by the Markov chain with probability:

$$min\left\{1, \frac{p(G'|D)}{p(G|D)}\right\}$$

Since $p(G|D) \propto p(D|G)p(G)$, the previous ratio corresponds to the Bayes factor of the two models which only involves efficient local computations. If the Markov chain is irreducible, i.e. there is a positive probability of reaching any DAG from any other DAG in the search space, then the process will converge to the posterior distribution $p(G|D)$.

In this setting, the randomly chosen candidate belongs usually to a NR or AR neighbourhood. We have extended this standard method for proposing a new candidate by letting the method to choose in a CR, NCR, RCARR and RCARNR neighbourhoods.

## 4 Experimental comparison

Here we present our experimental results showing that the new traversal operator in the space of DAGs brings much better results in reasonable time. For all the experiments we have used the BDe metric from (Heckerman *et al.* 1995) which is known to be invariant to equivalent DAGs (Chickering 1995). We have used an uniform prior distribution over the joint space of the parameters, as in Buntine (1991), and an equivalent sample size of 1.

## 4.1 The Alarm dataset

Throughout all the experimentation we have used the Alarm dataset from Herskovits (1991), which is a standard benchmark dataset to assess the performance of learning algorithms for Bayesian Networks. This dataset contains originally 20000 cases, from which the first 10000 where used by Cooper and Herskovits (1992) to assess their algorithm. From these first 10000 cases we have sampled six different datasets of two different sizes: three of 1000 cases and three of 5000 cases. The experiments reported on 10000 cases regard only the single dataset of the first 10000 cases. The dataset was synthetically generated from a Bayesian Network, introduced by Beinlich et al. (1989), of 37 nodes and 46 arcs. One of the arcs of this network is actually not supported by the data, as reported by Cooper and Herskovits (1992).

Table 1: Heuristic search. Performance is averaged over RCARR and RCARNR.

| smpl | rcar | performance | | score | | struct diff | |
|------|------|-------|--------|--------|--------|--------|--------|
|      |      | steps | sec/st | RCARNR | RCARR | RCARNR | RCARR |
| 1ka  | 0    | 55    | 0.27   | -11480.47 | -11480.47 | 29 | 29 |
|      | 2    | 58    | 0.40   | -11491.52±15.12 | -11470.46±15.79 | 18.90±3.06 | 16.50±2.00 |
|      | 4    | 55    | 0.44   | -11484.69±19.29 | -11473.41±14.18 | 18.00±3.28 | 16.30±2.18 |
|      | 7    | 54    | 0.43   | -11469.06±07.88 | -11470.75±14.67 | 15.50±1.55 | 15.60±1.88 |
|      | 10   | 53    | 0.43   | -11470.43±15.94 | -11464.03±11.00 | 14.80±2.15 | 15.20±1.78 |
| 1kb  | 0    | 60    | 0.28   | -11115.13 | -11115.13 | 28 | 28 |
|      | 2    | 58    | 0.40   | -11113.50±19.07 | -11105.10±20.62 | 18.10±2.77 | 14.70±3.87 |
|      | 4    | 56    | 0.42   | -11121.49±24.64 | -11090.15±08.51 | 17.90±4.41 | 11.10±2.35 |
|      | 7    | 53    | 0.42   | -11095.19±12.38 | -11083.13±05.07 | 13.40±2.55 | 10.00±1.47 |
|      | 10   | 53    | 0.43   | -11095.87±11.25 | -11094.17±18.72 | 12.40±2.05 | 11.50±2.11 |
| 1kc  | 0    | 62    | 0.61   | -11530.80 | -11530.80 | 37 | 37 |
|      | 2    | 60    | 0.41   | -11451.58±14.23 | -11453.70±15.75 | 18.20±2.23 | 15.90±3.10 |
|      | 4    | 59    | 0.43   | -11438.31±08.01 | -11436.65±07.46 | 14.90±2.95 | 13.40±1.94 |
|      | 7    | 56    | 0.67   | -11431.02±05.23 | -11427.84±03.62 | 11.80±1.61 | 10.80±0.81 |
|      | 10   | 53    | 0.86   | -11440.88±10.78 | -11428.89±08.95 | 13.70±2.13 | 11.00±1.17 |
| 5ka  | 0    | 69    | 1.54   | -55249.43 | -55249.43 | 46 | 46 |
|      | 2    | 66    | 2.50   | -55072.99±67.61 | -54993.41±08.70 | 11.60±6.70 | 7.20±2.23 |
|      | 4    | 57    | 2.09   | -55051.93±40.93 | -54992.40±10.92 | 7.90±2.12 | 5.20±1.38 |
|      | 7    | 56    | 2.14   | -55024.53±48.12 | -54989.70±10.12 | 7.10±2.17 | 4.90±1.37 |
|      | 10   | 56    | 2.08   | -55025.19±43.49 | -54985.99±06.68 | 6.10±1.95 | 5.10±2.49 |
| 5kb  | 0    | 57    | 0.92   | -54732.19 | -54732 | 33 | 33 |
|      | 2    | 57    | 2.02   | -54679.46±34.06 | -54641.27±60.60 | 12.50±6.25 | 6.10±4.08 |
|      | 4    | 56    | 1.35   | -54610.82±15.24 | -54607.60±14.34 | 5.20±3.93 | 3.80±1.38 |
|      | 7    | 53    | 1.29   | -54611.85±25.63 | -54602.77±10.93 | 4.50±1.52 | 3.70±1.31 |
|      | 10   | 52    | 1.28   | -54602.98±10.85 | -54606.47±12.48 | 4.00±1.26 | 4.10±1.32 |
| 5kc  | 0    | 59    | 0.88   | -54454.16 | -54454.16 | 36 | 36 |
|      | 2    | 63    | 1.19   | -54340.02±16.48 | -54335.27±32.25 | 10.20±3.97 | 8.00±2.18 |
|      | 4    | 59    | 1.20   | -54335.49±19.99 | -54326.25±11.15 | 8.60±1.91 | 8.40±2.05 |
|      | 7    | 55    | 1.25   | -54331.19±12.09 | -54315.06±07.33 | 8.50±1.55 | 7.70±1.35 |
|      | 10   | 55    | 1.28   | -54363.17±52.63 | -54329.40±11.13 | 10.00±2.18 | 8.50±1.85 |
| 10k  | 0    | 56    | 1.86   | -108697.78 | -108697.78 | 21 | 21 |
|      | 2    | 56    | 2.23   | -108495.65±68.33 | -108463.65±46.17 | 4.90±2.20 | 5.40±4.10 |
|      | 4    | 54    | 2.28   | -108549.53±63.63 | -108437.83±35.72 | 6.80±2.25 | 1.60±0.90 |
|      | 7    | 50    | 2.29   | -108477.50±52.06 | -108485.55±58.14 | 5.50±3.22 | 2.80±1.11 |
|      | 10   | 50    | 2.41   | -108468.56±53.07 | -108477.98±51.65 | 4.20±1.34 | 3.30±1.17 |

## 4.2 Heuristic learning

Heuristic learning is assessed as follows. On each of the seven datasets the HCMC was run ten times for four different cardinalities of RCAR (2,4,7 and 10) and three different neighbourhoods (AR, RCARR and RCARNR).

The traditional hill-climber that uses an AR neighbourhood will be referred here as RCAR 0. The reason of running the HCMC several times is obvious since this new hill-climber performs random moves that may lead to different results in different runs. The maximum number of trials for escaping local maxima was set to 50. The results have been averaged over ten runs and confidence intervals, at a level of 95% for the means of score and structural difference, have been included. We may see these results in table 1.

More concretely, these are the four different measures that have been taken:

- steps: number of steps ($g = g'$ in the algorithm) of the HCMC. The lower the faster.
- sec/st: speed of the HCMC in seconds per step.
- score: score of the learned Bayesian Network.
- struct diff: structural difference between the essential graph of the original alarm network and the essential graph of the learned network.

As we may appreciate, there is a substantial difference in using RCAR within the hill-climber. The best performance is achieved when a NCR neighbourhood is also used. The most striking evidence lies in the case of 10000 cases and RCARR. There, an average of just only 1.6 structural differences is achieved in about 54 steps. We should remark that on *eight* out of ten of those runs, there was only just one structural difference, corresponding to the missing arc not supported by the data, which makes the confidence interval very tight. In some of those eight times, the result was reached in 49 steps and the same result was reached for RCAR 7 and 10 even in 48 steps, which is extremely close to the optimal path of the right result (46 additions).

## 4.3 MCMC learning

The goal of the experimentation with MCMC is twofold. In the first place, to show that RCAR affords a higher mobility of the Markov chain and a faster convergence. In the second place, to confirm the results in heuristic learning and shed some light over the reasons behind the success of the use of RCAR. For all the previously described datasets, the Markov chain started from the empty graph. For the dataset of 10000 cases, we also ran the chain starting from the true Alarm network with one missing arc not supported by the data.

As described previously, the Markov chain moves among the search space of DAGs. Nevertheless, when the Markov chain is finished, the posterior distribution of DAGs, and other derived quantities indexed by DAGs, are transformed into quantities or distributions related to essential graphs. For that purpose, we have used the algorithm of Chickering (1995) that obtains the corresponding essential graph of a given DAG.

Table 2: Mobility of the chain and Kullback-Leibler distance per essential graph

|  | size | AR | CR | NCR | RCARR | RCARNR |
|---|---|---|---|---|---|---|
| number essential graphs | 1k | 1764 | 1622 | 1632 | 1898 | 2017 |
|  | 5k | 830 | 780 | 660 | 991 | 955 |
|  | 10k | 561 | 470 | 526 | 727 | 654 |
|  | 10k* | 553 | 485 | 612 | 577 | 626 |
| K-L per e.g. | 1k | 4.36525 | 4.46228 | 4.37798 | 4.33639 | 4.38295 |
|  | 5k | 3.78184 | 3.78113 | 3.95602 | 3.66680 | 3.75692 |
|  | 10k | 3.73537 | 4.05203 | 3.97652 | 3.53715 | 3.75776 |
|  | 10k* | 2.70025 | 2.70035 | 2.70018 | 2.70029 | 2.70046 |

The mobility of the chain is an important aspect because the higher the mobility, the lower the chance that the approximated posterior distribution does not reflect an important area of the search space. In table 2 we may see the averages across the samples and across the RCAR cardinalities of 2,4 and 10 of the different essential graphs visited during the process, as well as the average Kullback-Leibler distance per essential graph. It is clear that RCAR yields a higher mobility of the chain since more essential graphs were visited when RCAR was used.

This higher mobility results in a better choice of the DAGs during the process, as we may see from the lower K-L ratios for the cases where RCAR was used. The asterisk in table 2 denotes the case where the *almost* true network was used as starting point. In this latter situation, the K-L ratio does not show a gain while using RCAR. This is because the chain starts from a good point and all the neighbouring models still provide a good K-L distance.

In order to show how RCAR improves the convergence of MCMC we will use three diagnostics introduced by Giudici and Castelo (2000) in the context of Bayesian and Decomposable Networks. The first one is the average number of edges along the iterations. It is expected that this average gets close to the number of edges of the true DAG as the chain approaches convergence.

The second diagnostic is the approximation of the marginal of the data $p(D) = \sum_G p(G, D)$ along the iterations. This marginal is approximated as $\hat{p}(D) = p(G, D)/\hat{p}(G|D)$ by averaging over the five DAGs with highest posterior at the $i$th iteration, such that the higher is the better. We may see these two diagnostics for the case of 10000 records starting from the empty DAG and using 4 as a parameter for RCAR in figure 2. The plots show that RCARR4 is able to approach faster both the optimal number of edges and a higher marginal of the data $p(D)$.

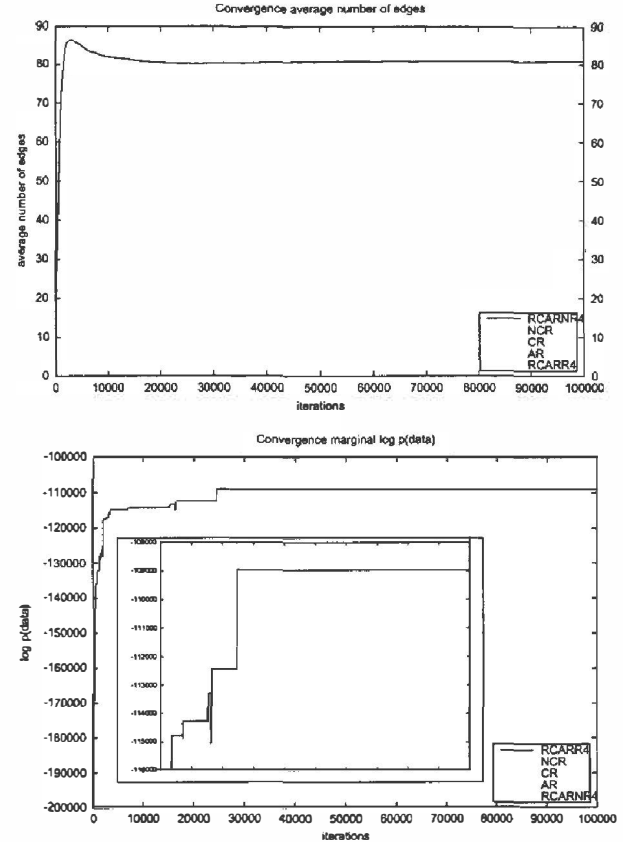

Figure 2: Convergence diagnostics for the 10k dataset. Legend is ordered with lines.

The third convergence diagnostic is the plot of the posterior distribution of the different numbers of edges. This distribution tends to be normal and centered close to the cardinality of the graph for which the Markov chain gives the highest posterior. In figure 3 we show this diagnostic for the 10000 case dataset, starting from the almost true Alarm network (noted with an asterisk in the legend), and from the empty network. We may see that only RCARR is able to reproduce quite similar posteriors in either starting point.

Finally we look at the amount of members of each equivalence class of DAGs visited during the process. For each different essential graph obtained from the transformation of the DAGs, we computed a lower bound on the cardinality of the class it represents. This lower bound is computed as follows. For each connected component $i$ of the essential graph, let $q_i$ be the number of *reversible* edges. The lower bound on the number of members of the equivalence class represented by the essential graph is defined as $\prod_i (q_i + 1)$.

In figure 4 we can see the comparison of the lower bound with respect to the number estimated from the MCMC process. We have taken the picture cor-responding to the 10000 cases dataset for which the Markov chain started from the almost true Alarm network. On the x-axis we find the essential graphs ordered by the moment in time the first DAG member was visited by the chain. Since the plots for the CR and NCR neighbourhoods are similar to the one of AR, and the same happens for RCARNR and RCARR, we only compare here AR with RCARR.

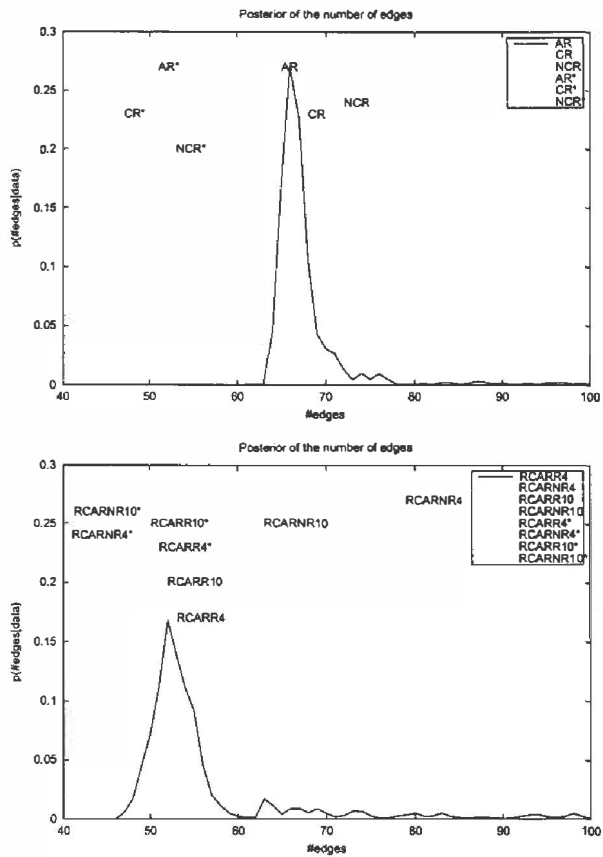
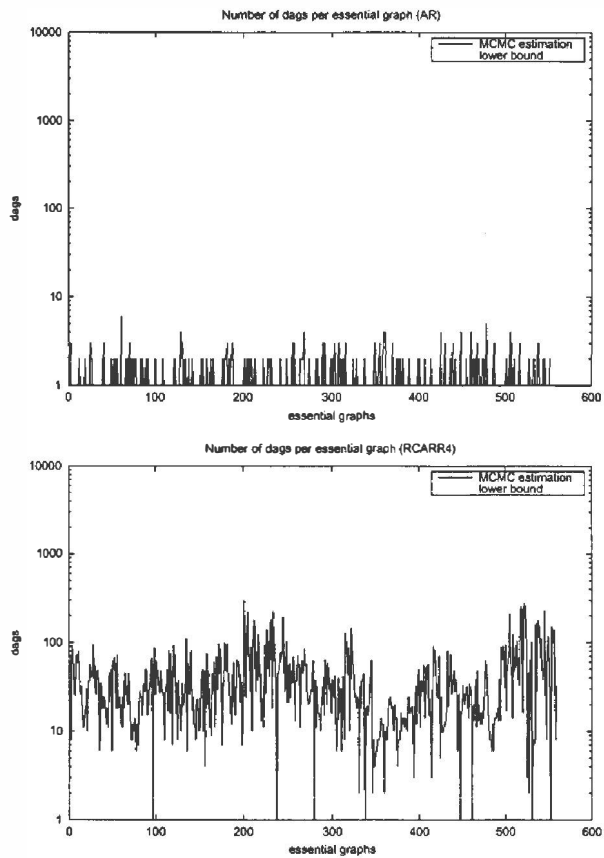

Figure 3: Comparison convergence ability.

Figure 4: DAGs per essential graph.

As it follows from these plots, RCAR approaches better the shape of the different cardinalities of the classes. The better estimation of RCAR is the empirical evidence from which follows that RCAR is able to reach more neighbouring equivalence classes of DAGs, which is the ultimate reason for learning faster and better Bayesian Networks.

Table 3: Performance MCMC

|         | AR       | RCARR4   | RCARR10  | AR*      | RCARR4*  | RCARR10* |
|---------|----------|----------|----------|----------|----------|----------|
| iter/sec | 71.3    | 25.4     | 39.0     | 100.4    | 60.3     | 55.3     |
| acc/rej | 8.54e-03 | 8.12e-03 | 7.87e-03 | 7.95e-03 | 5.87e-03 | 6.15e-03 |

In table 3 we can see the extra computational cost of RCAR. The first row contains the accumulated average number of iterations per second at the end of the run of the Markov chain. The second row contains the accepts/rejects ratio where a lower ratio speeds up the process.

For clarity we only report comparison between AR, RCARR4 and RCARR10, as the differences are similar for other combinations. This comparison has been done for the 10000 cases dataset starting from the empty and the true Alarm network (noted with an asterisk). We may see that in any of the two cases, using RCAR is between two and three times slower than not using it. This cost agrees with the one we observed for heuristic learning and we consider it to be a very good trade-off.

## 5  Conclusion

The standard AR neighbourhood for a given DAG is not invariant to other DAGs within the same equivalence class. In order to solve this problem, we have introduced new concepts of traversal operators for DAGs, based upon the idea to make more moves within the equivalence classes by applying the covered arc reversal operation.

The experiments with RCAR within the MCMC method show a higher mobility of the chain and a faster convergence rate according to the diagnostics, without compromising the already heavy computational cost of the method. Madigan et al. (1996) point out a problem for Markov chains that do not move in the space of essential graphs. The approximated posterior may be proportional to the size of the equivalence classes. Obviously RCAR does not solve this problem but there is no guarantee that the moves in the space of essential graphs approximate the posterior well. We believe that the quality of the approximation depends more on the concept of neighbourhood than on the search space used.

We have shown usage of RCAR within the hill climbing heuristic where it brings very impressive results in reasonable time as, e.g. recovering 45 out of the 46 edges of the Alarm network in 48 steps (single local transformations). Our work included a few user defined constants (number of covered arc reversals in RCAR, number of repetitions of RCAR in the local maximum of the hill-climber) which could be assigned in a more clever way. We expect that it could bring another improvement of obtained results.

Moreover we have shown that the non-covered arc reversal operation, as used in RCARR, is beneficial for heuristic learning and MCMC. This could seem counterintuitive to Meek's conjecture. However in MCMC, longer runs of the Markov chain ($10^6$ it.) starting from the true Alarm network, which we do not report here because of lack of space, show that RCARNR converges faster to the probability distribution where the true Alarm network gets the highest posterior. This suggests that some mixed strategy of both, RCARR and RCARNR, would probably afford the fastest convergence. As a closing remark, we have demonstrated the practical importance of the highly theoretical problem of Bayesian Networks inclusion as the general Meek's conjecture remains still open.

### Acknowledgements

We thank Chris Meek for making us familiar with his conjecture, Steven Gillispie and Bertran Steinsky for keeping us up to date on their discoveries, Paolo Giudici for his help with the MCMC method and Arno Siebes and the anonymous reviewers for useful remarks. This research has been supported by grant FRVŠ n. 2001/1433. The authors benefited from participation in the seminar "Conditional independence structures" (Fields Institute –Toronto, October 1999).

### References

I. Beinlich, H. Suermondt, R. Chavez and G. Cooper (1989) The ALARM monitoring system, in Proc. 2nd European Conf. on AI and Medicine, pp. 247–256

W. Buntine (1991) Theory refinement on Bayesian Networks, in Proc. UAI (B. D'Ambrosio, P. Smets and P. Bonissone eds.), M. Kaufmann, pp. 52–60.

M. D. Chickering (1995) A transformational characterization of equivalent Bayesian Networks, in Proc. UAI (P. Besnard, S. Hanks eds.), M. Kaufmann, pp. 87–98.

M. D. Chickering (1996) Learning equivalence classes of Bayesian Network structures, in Proc. UAI (E. Horvitz, F. Jensen eds.), M. Kaufmann, pp. 150–157.

G. Cooper and E.H. Herskovits (1992): A Bayesian method for the induction of probabilistic networks from data, *Machine Learning*, 9, pp. 309–405.

M. Frydenberg (1990) The chain graph Markov property, *Scand. J. of Statistics*, 17, pp. 333–353.

S. Gillispie and M. Perlman (2001) Enumerating Markov Equivalence Classes of Acyclic Digraph Models, in UAI (J. Breese, D. Koller eds), M. Kaufmann.

P. Giudici and R. Castelo (2000) Improving Markov Chain Monte Carlo model search for Data Mining. Technical report 117. University of Pavia, Italy.

D. Heckerman, D. Geiger and D.M. Chickering (1995) Learning Bayesian Networks: the combination of knowledge and statistical data, *Machine Learning*, 20, pp. 197–243.

E.H. Herskovits (1991) Computer-Based Probabilistic-Network Construction. Doctoral dissertation. Medical Information Sciences. Stanford University.

T. Kočka, R. Bouckaert and M. Studený (2001) On characterizing inclusion of Bayesian Networks, in Proc. UAI (J. Breese, D. Koller eds.), M. Kaufmann.

D. Madigan and J. York (1995) Bayesian graphical models for discrete data, *Int. Stat. Rev.*, 63, 215-232.

D. Madigan, S. Andersson, M. Perlman and C. Volinsky (1996) Bayesian model averaging and model selection for markov equivalence classes of acyclic digraphs. *Comm. in Stat. (th. and meth.)*, 25 (11): 2493–2512.

C. Meek (1997) Graphical models, selecting causal and statistical models, PhD thesis, CMU.

R.W. Robinson (1973) Counting labeled accyclic digraphs in New Directions in the Theory of Graphs (ed. Harary F.) Academic Press, New York, p. 239-273.

B. Steinsky (2000) Enumeration of Labelled Chain Graphs and Labelled Essential Directed Acyclic Graphs, University of Salzburg, Austria.

T. Verma and J. Pearl (1991) Equivalence and synthesis of causal models, in Proc UAI (P. Bonissone, M. Henrion, L. Kanal, J. Lemmer eds.), Elsevier, pp. 220-227.